\documentclass{ieeeaccess}
\usepackage{cite}
\usepackage{amsmath,amssymb,amsfonts}
\usepackage{algorithmic}
\usepackage{graphicx}
\usepackage{textcomp}
\usepackage{url}
\usepackage{hyperref}   
\usepackage[table]{xcolor}  
\hypersetup{
    colorlinks=true,   
    linkcolor=blue,    
    citecolor=blue,    
    urlcolor=blue,     
    filecolor=blue,    
    pdfborder={0 0 0}  
}
\usepackage{soul}
\usepackage{xcolor}
\usepackage[ruled,vlined]{algorithm2e}
\sethlcolor{yellow}
\usepackage{placeins} 

\usepackage{bm}
\makeatletter
\AtBeginDocument{\DeclareMathVersion{bold}
\SetSymbolFont{operators}{bold}{T1}{times}{b}{n}
\SetSymbolFont{NewLetters}{bold}{T1}{times}{b}{it}
\SetMathAlphabet{\mathrm}{bold}{T1}{times}{b}{n}
\SetMathAlphabet{\mathit}{bold}{T1}{times}{b}{it}
\SetMathAlphabet{\mathbf}{bold}{T1}{times}{b}{n}
\SetMathAlphabet{\mathtt}{bold}{OT1}{pcr}{b}{n}
\SetSymbolFont{symbols}{bold}{OMS}{cmsy}{b}{n}
\renewcommand\boldmath{\@nomath\boldmath\mathversion{bold}}}
\makeatother

\def\BibTeX{{\rm B\kern-.05em{\sc i\kern-.025em b}\kern-.08em
    T\kern-.1667em\lower.7ex\hbox{E}\kern-.125emX}}

\begin{document}
\history{Date of publication xxxx 00, 0000, date of current version xxxx 00, 0000.}
\doi{10.1109/ACCESS.2024.0429000}

\title{Measurement of Medial Elbow Joint Space using Landmark Detection}

\author{
\uppercase{Shizuka Akahori}\href{https://orcid.org/0009-0002-0448-0497}{\includegraphics[scale=0.07]{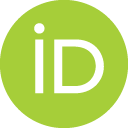}}\authorrefmark{1}, 
\uppercase{Shotaro Teruya}\href{https://orcid.org/0009-0006-4851-0837}{\includegraphics[scale=0.07]{ORCIDiD_icon128x128.png}}\authorrefmark{2}, 
\uppercase{Pragyan Shrestha}\href{https://orcid.org/0009-0000-8276-4116}{\includegraphics[scale=0.07]{ORCIDiD_icon128x128.png}}\authorrefmark{1}, 
\uppercase{Yuichi Yoshii}\href{https://orcid.org/0000-0003-1447-4664}{\includegraphics[scale=0.07]{ORCIDiD_icon128x128.png}}\authorrefmark{3}, 
\uppercase{Ryuhei Michinobu}\authorrefmark{4}, 
\uppercase{Satoshi Iizuka}\href{https://orcid.org/0000-0001-9136-8297}{\includegraphics[scale=0.07]{ORCIDiD_icon128x128.png}}\authorrefmark{5}, 
\uppercase{Akira Ikumi}\href{https://orcid.org/0000-0001-6034-1593}{\includegraphics[scale=0.07]{ORCIDiD_icon128x128.png}}\authorrefmark{2}, 
\uppercase{Hiromitsu Tsuge}\authorrefmark{6}, 
and \uppercase{Itaru Kitahara}\href{https://orcid.org/0000-0002-5186-789X}{\includegraphics[scale=0.07]{ORCIDiD_icon128x128.png}}\authorrefmark{7} \IEEEmembership{Member, IEEE}
}

\address[1]{Graduate School of Science and Technology, University of Tsukuba, Tennodai 1-1-1, Tsukuba, Ibaraki 305-8571, Japan}
\address[2]{Department of Orthopaedic Surgery, University of Tsukuba, Tennodai 1-1-1, Tsukuba, Ibaraki 305-8571, Japan}
\address[3]{Department of Orthopaedic Surgery, Ibaraki Medical Center, Tokyo Medical University, Ami 3-20-1, Inashiki, Ibaraki 300-0332, Japan}
\address[4]{Tsukuba Wellness Orthopaedics, Sasagi 2011-54, Tsukuba, Ibaraki 305-0043, Japan}
\address[5]{Institute of Systems and Information Engineering, University of Tsukuba, Tennodai 1-1-1, Tsukuba, Ibaraki 305-8571, Japan}
\address[6]{Department of Orthopaedic Surgery, Kikkoman General Hospital, Miyazaki 100, Noda, Chiba 278-0005, Japan}
\address[7]{Center for Computational Sciences, University of Tsukuba, Tennodai 1-1-1, Tsukuba, Ibaraki 305-8571, Japan; 
Center for Cyber Medicine Research, University of Tsukuba, Tennodai 1-1-1, Tsukuba, Ibaraki 305-8575, Japan}

\tfootnote{
This work was supported by Japan Science and Technology Agency (JST) through Grant Number JPMJFS2106, JSPS KAKENHI Grant Number 24K14397, and Multidisciplinary Cooperative Research Program in CCS, University of Tsukuba.
This work involved human subjects in its research. Approval of all ethical and experimental procedures and protocols was granted by the Ethics Committee in the Faculty of Medicine, University of Tsukuba (October 7, 2022, No. 1517-3) and the Ethics Committee of the Center for Computational Science, University of Tsukuba (December 11, 2023, No. 23-004; Jun. 24, 2024, No. 24-005), and performed in line with the principles of the Declaration of Helsinki. Informed consent was obtained from all individual participants included in the study.}

\markboth
{Author \headeretal: Preparation of Papers for IEEE TRANSACTIONS and JOURNALS}
{Author \headeretal: Preparation of Papers for IEEE TRANSACTIONS and JOURNALS}

\corresp{Corresponding author: Shizuka Akahori (e-mail: akahori.shizuka@image.iit.tsukuba.ac.jp).}

\begin{abstract}
Ultrasound imaging of the medial elbow is crucial for early detection of Ulnar Collateral Ligament (UCL) injuries. Specifically, measuring the elbow joint space on ultrasound images enables the assessment of valgus instability. Automating this measurement requires a precisely annotated dataset to leverage recent advances in deep neural networks; however, no publicly available dataset has been proposed to date. In this work, we introduce a novel medial elbow ultrasound dataset with landmark annotations, facilitating the development of deep neural networks for joint space measurement. The dataset comprises 4,109 medial elbow ultrasound images from 22 participants, with precise landmark annotations on the humerus and ulna, performed under the supervision of three orthopedic surgeons. We evaluate several landmark detection models on our dataset and demonstrate its effectiveness in training models. Additionally, we propose Shape Subspace Refinement (SSR) to refine landmark heatmap peaks, improving the accuracy of heatmap-based landmark detection models. Furthermore, we show that the detected landmarks can be used for point-based segmentation of the humerus and ulna. The dataset is publicly available at \url{https://github.com/Akahori000/Ultrasound-Medial-Elbow-Dataset}.
\end{abstract}


\titlepgskip=-21pt

\maketitle

\begin{figure*}[t!]
    \centering
    \includegraphics[width=0.95\textwidth]{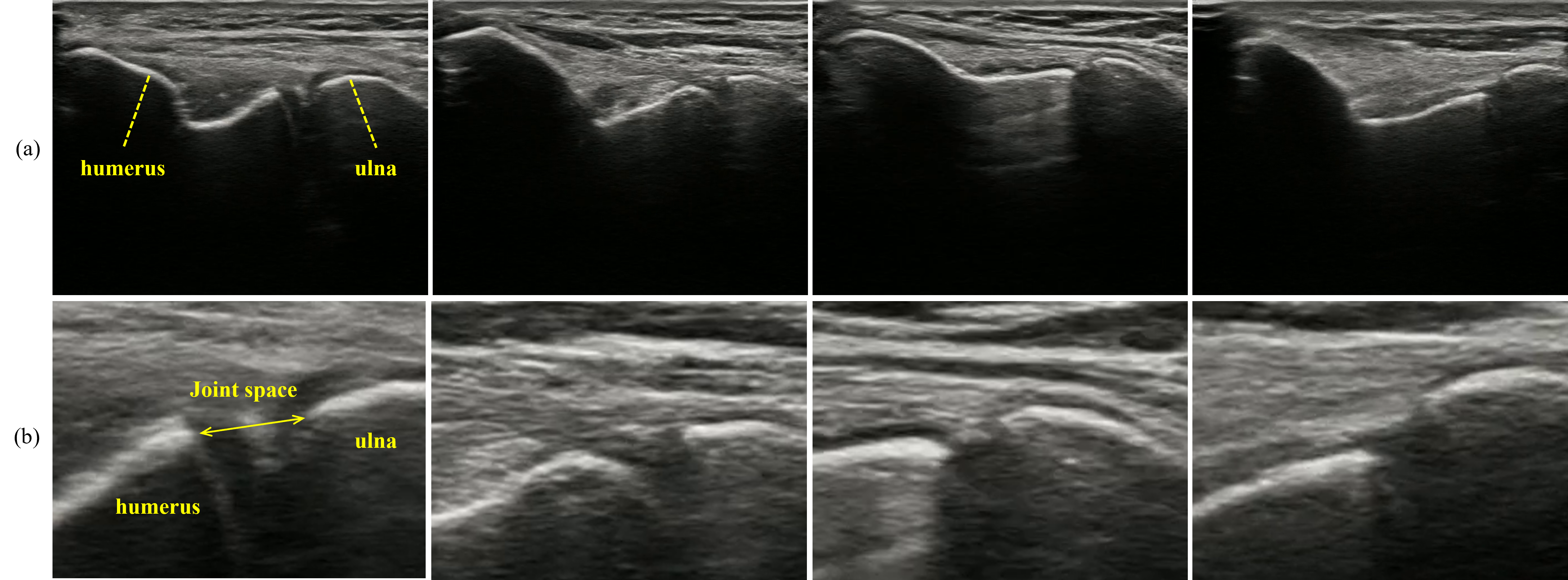}
    \caption{\textbf{(a) The ultrasound images of the medial elbow obtained from different participants. (b) The enlarged views of the joint space positions. The left and right hyperechoic lines located in the middle of each image are the humerus and ulna, respectively. The space between humerus and ulna is the joint space to be measured.}\label{fig_images}}
\end{figure*}

\section{Introduction}
\label{sec:introduction}

\PARstart{U}{ltrasound} imaging is an accessible, cost-effective, and noninvasive diagnostic technique widely used in clinical practice. In orthopedics, medial elbow ultrasound screening is commonly performed in athletes. Specifically, measuring the ulnohumeral joint space using ultrasound images provides critical insights into the risk of Ulnar Collateral Ligament (UCL) injury \cite{bib_gap2, bib_gap1}. An increase in the joint space under stress conditions suggests joint laxity and potential UCL damage of the elbow. Shanley et al. \cite{bib_gap1} measured joint space by defining landmarks on the edges of the ulna and humerus in ultrasound images of the medial elbow. Their study showed that pitchers with UCL injuries exhibit significantly greater joint space compared to those without injuries ($6.5\pm1.2$ mm vs. $5.3\pm1.2$ mm).  
To effectively measure joint laxity, Michinobu et al. \cite{bib_gap3} investigated the optimal elbow positioning during ultrasonography.

Several studies \cite{bib_elbow1, bib_elbow2, bib_elbow3} have explored deep learning-based diagnosis using elbow ultrasound images. These studies have focused on osteochondritis dissecans or median nerve detection.
However, none of the studies have discussed the automation of joint space measurements for diagnosing UCL injuries. Although precise and consistent measurement is required, it is challenging due to anatomical variability, noise, and artifacts.
Fig.~\ref{fig_images} illustrates ultrasound images of the medial elbow from different participants (a) and the enlarged views of the humerus and ulna near the joint space (b). These images show that bone shapes vary among individuals, and bone edges are often indistinct. Also, hyperechoic lines and artifacts often appear near bone surfaces or within the joint space.

To address these challenges, training a model on a dataset of diverse joint images with accurate annotations is needed.
However, no ultrasound dataset exists to measure the joint space in the medial elbow, and developing such a dataset is complex and demanding. Data acquisition requires advanced technical skills, adherence to ethical standards, and careful consideration for participants' consent and privacy. Also, annotation requires specialized knowledge.

In this study, we introduce a medial elbow ultrasound dataset consisting of 4,109 images from 22 individuals. The dataset contains the precise landmark annotation on the humerus and ulna based on the expertise of three orthopedic surgeons. We evaluated joint space measurement methods on our proposed dataset using several landmark detection approaches. Although heatmap-based landmark detection methods generally achieve high accuracy, they may occasionally produce multiple peaks on a heatmap, leading to incorrect detection. To mitigate this, we propose Shape Subspace landmark Refinement (SSR) by measuring the geometric differences between detected and reference landmark positions. 
The results show that our measurement method is sufficiently accurate to diagnose UCL injuries.
Finally, we demonstrate point-based segmentation of the humerus and ulna using detected landmarks as inputs, showing an application of our landmark detection in the medial elbow. 

\begin{figure*}[t!]
    \centering
    \includegraphics[width=0.9\textwidth]{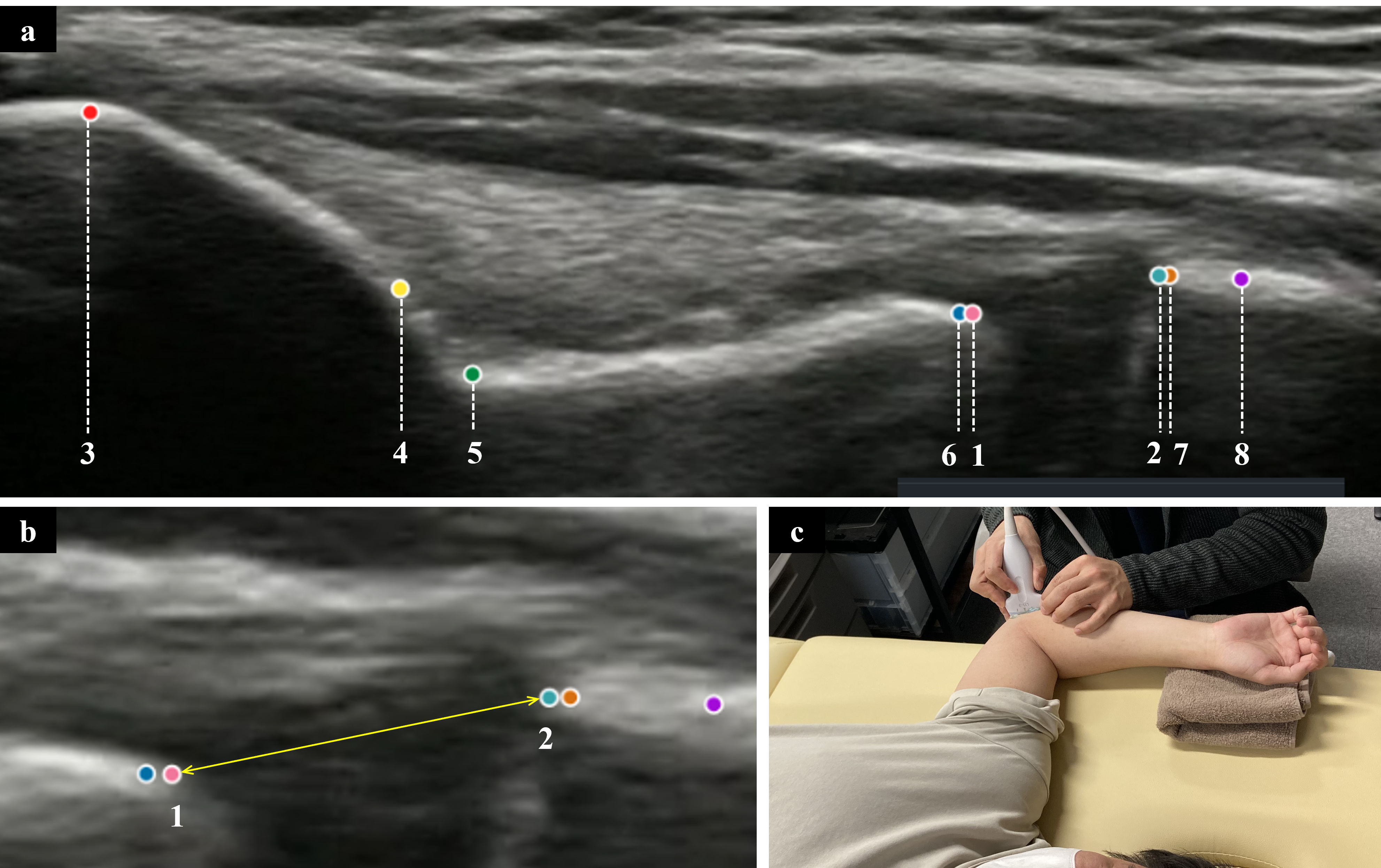}
    \caption{\textbf{(a) The annotated landmarks: 1) Distal end of humeral trochlea, 2) Proximal end of ulnar coronoid process. Landmarks 1 and 2 are the joint space measurement points.
3) Tip of humeral medial epicondyle, 4) Deep end of the intersection of humeral medial epicondyle and ulnar collateral ligament, 5) Distal end of humeral medial epicondyle, 6) Proximal point adjacent to landmark 1 on the center of hyperechoic line of humeral trochlea, 7) Distal point adjacent to landmark 2 on the center of hyperechoic line of ulnar coronoid process, 8) Distal point of ulnar coronoid process. The white line on which landmarks 1 and 3 through 6 are located represents the humerus, while the white line on which landmarks 2, 7, and 8 are located represents the ulna.
(b) The enlarged view of joint space. The yellow arrow is the joint space length to be measured. Landmarks 1 and 2 are the joint space measurement points. (c) The participant's posture during the examination.}\label{fig_annotations}}
\end{figure*}

\section{Related work}
\subsection{Landmark detection Methods}
Landmark detection, also known as keypoint detection, is a method of locating keypoints in images. Various methods have been proposed for pose estimation.

Regression-based approaches \cite{bib_regression1,bib_regression2,bib_regression3,bib_regression5,bib_regression6,bib_yolo7,bib_yolov8} directly predict landmark coordinates.
These methods are simple and computationally efficient, but often lack precision, especially in complex poses.
Heatmap-based methods overcome this by generating heatmaps where the intensity represents the likelihood of a landmark's location, achieving better spatial accuracy. 
Numerous studies have leveraged Convolutional Neural Networks (CNNs) for this purpose \cite{bib_CNN1,bib_CNN2,bib_CNN3,bib_CNN4,bib_CNN5,bib_CNN6,bib_CNN7,bib_CNN8,bib_CNN9,bib_CNN10, bib_UNet, bib_HRNet}. Among these, U-Net \cite{bib_UNet} and its extensions are widely used in localization tasks of medical images. The architecture utilizes skip connections to copy the feature maps from each stage of the encoder and directly pass them to the decoder's corresponding stage.
HRNet \cite{bib_HRNet} maintains high-resolution representations throughout the whole process, allowing for accurate localization with multi-resolution features.
Recently, Transformer-based heatmap models \cite{bib_vit1,bib_transpose,bib_hrformer,bib_vitpose} have been introduced, leveraging self-attention to model long-range dependencies. ViTPose \cite{bib_vitpose} exemplifies this with a simple Vision Transformer architecture. 
Lastly, token-based methods \cite{bib_tokenpose, bib_pct} learn the relationships between skeletal joints using tokens, where 
PCT \cite{bib_pct} designs each token to represent a substructure of joints.

Several landmark detection approaches have been applied to ultrasound images to measure anatomical structures \cite{bib_ultra1, bib_ultra2, bib_ultra4, bib_ultra5, bib_ultra7}. U-Net and its extensions are used to measure left ventricle dimensions \cite{bib_ultra2, bib_ultra5}. Mask R-CNN-based architecture has been employed to evaluate developmental dysplasia of the hip from infant ultrasound images \cite{bib_ultra4}. HRNet-based regression architecture is proposed to assess infant growth \cite{bib_ultra7}.

\subsection{Shape Subspace}
Shape Subspace is a method to represent the 3D points of a 3D structure as a linear subspace within a high-dimensional vector space. By measuring the similarity between two shape subspaces, the geometrical difference of 3D structures can be measured \cite{Igarashi}. Shape subspace has been utilized to classify 3D objects and human faces \cite{yoshimura, yataka}.

\subsection{Point-based Segmentation}
Segment Anything in Images and Videos (SAM2) \cite{bib_sam2} is designed for interactive segmentation tasks, capable of processing images and videos. Using point or bounding-box prompts, SAM2 can generate accurate segmentation masks. Its robust performance is driven by a large-scale training dataset and a transformer-based architecture.

\section{Dataset Construction}
\subsection{Data Collection}\label{datasetcollection}  
We constructed an ultrasound medial elbow dataset of 4,109 images from 22 participants (ages 22–56, men and women, from five countries). All participants had no symptoms related to the elbow joint. The images were acquired using musculoskeletal ultrasonography (SONIMAGE MX1, KONICA MINOLTA, Tokyo, Japan) with an 11-MHz linear probe by two experienced orthopedic surgeons.
The participants lay on a bed with their elbows flexed at 90 degrees as in \cite{bib_gap3}. Then, a surgeon moved the gel-applied probe around the medial elbow, as shown in Fig.~\ref{fig_annotations}(c). The images capture the medial epicondyle, UCL, medial surface of the humeral trochlea, and coronoid process of the ulna, a key region for diagnosing medial elbow conditions \cite{bib_gap3}. This procedure is the same as that used in routine medial elbow examinations (e.g., for baseball players).
The image resolution is 528 × 672 pixels, corresponding to 0.0567 mm per pixel. In the images, the humerus appears on the left and the ulna on the right. Fig.~\ref{fig_images} shows sample images of different participants with the enlarged joint space views. Representative images and additional information for each participant are provided in the supplementary material.

\begin{figure*}[t!]
\centering
\includegraphics[width=0.9\textwidth]{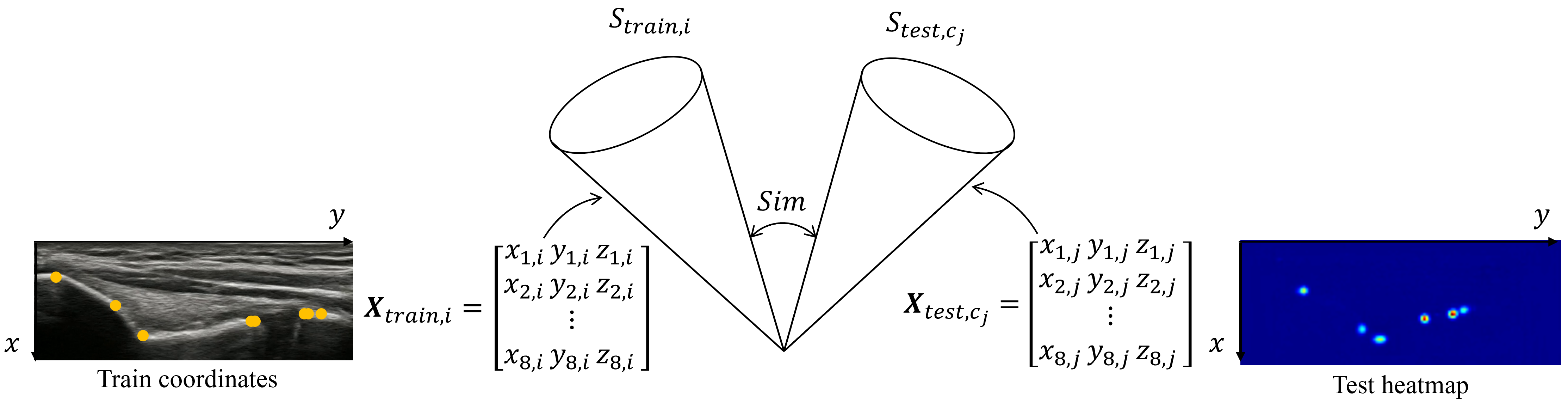}
\caption{Visualization of shape subspace refinement (SSR).
Shape matrices \(\mathbf{X}_{\text{train},i}\) and \(\mathbf{X}_{\text{test},c_j}\) are derived from the landmark coordinates of the training data and the test heatmaps, respectively. Shape subspaces \(\mathbf{S}_{\text{train},i}\) and \(\mathbf{S}_{\text{test},c_j}\) are computed from the shape matrices, and the similarity between the two shape subspaces is compared. Note that \(z\) in \(\mathbf{X}_{\text{train},i}\) and \(\mathbf{X}_{\text{test},c_j}\) is set to zero to generate shape subspace from 2D coordinates. }\label{fig_shape}
\end{figure*}





\subsection{Annotation}\label{Annotation}
The obtained 4,109 images were manually annotated with landmarks at sub-pixel levels by the authors under the supervision of three orthopedic surgeons. 
The landmarks include two joint space measurement points: the distal end of the humeral trochlea, and the proximal end of the ulnar coronoid process, corresponding to the positions described in \cite{bib_gap1}. In addition to the joint space measurement points, six additional points on the humerus and ulna were annotated. These points are of particular interest to sonographers, who primarily focus on bone structures during ultrasound image interpretation.
The annotated landmarks were meticulously reviewed and discussed by the three orthopedic experts, ensuring the reliability of the dataset.

Fig.~\ref{fig_annotations}(a) shows the annotated landmarks: 1) Distal end of humeral trochlea, 2) Proximal end of ulnar coronoid process. Landmarks 1 and 2 are the joint space measurement points.
3) Tip of humeral medial epicondyle, 4) Deep end of the intersection of humeral medial epicondyle and ulnar collateral ligament, 5) Distal end of humeral medial epicondyle, 6) Proximal point adjacent to landmark 1 on the center of hyperechoic line of humeral trochlea, 7) Distal point adjacent to landmark 2 on the center of hyperechoic line of ulnar coronoid process, 8) Distal point of ulnar coronoid process.
The yellow line segment between landmarks 1 and 2 in Fig.~\ref{fig_annotations}(b) shows the joint space to be measured. 

\subsection{Ethical Approval and Informed Consent}\label{Ethical}
This study was conducted in accordance with the principles of the Declaration of Helsinki. Ethical approval was obtained from the Ethics Committee of the Faculty of Medicine, University of Tsukuba (October 7, 2022, Approval No. 1517-3) and the Ethics Committee of the Center for Computational Science, University of Tsukuba (December 11, 2023, Approval No. 23-004; June 24, 2024, Approval No. 24-005). Informed consent was obtained from all participants involved in the study.

\section{Method}
\subsection{Landmark Detection Models}\label{landmarkdetection}
Several landmark detection models are compared: HRNet \cite{bib_HRNet}, ViTPose \cite{bib_vitpose}, and U-Net \cite{bib_UNet} as heatmap-based methods, YOLOv8 \cite{bib_yolov8} as a regression-based method, and PCT \cite{bib_pct} as a token-based method. 

\subsubsection{HRNet, ViTPose, and U-Net}  
The landmark detection process of HRNet, ViTPose, and U-Net is represented in \eqref{eq1}, where \(f\) is the detection model and \(\mathbf{x} \in \mathbb{R}^{W \times H \times 3}\) is the input image with width \(W\), height \(H\), and 3 channels. \(f(\mathbf{x})\) outputs heatmaps of size \(W' \times H' \times N_K\), where \(W'\) and \(H'\) are the heatmap width and height, and \(N_K\) is the number of landmarks.
Each heatmap corresponds to a specific landmark, meaning that \( N_K \) heatmaps are generated.
The \(\arg\max(f(\mathbf{x}))\) operation returns the coarse landmark coordinates \(\mathbf{p}' \in \mathbb{R}^{N_K \times 2}\). The Unbiased Data Processing (UDP) function \cite{bib_udp1, bib_CNN10} refines these to sub-pixel accuracy, yielding the final landmark positions \(\mathbf{p} \in \mathbb{R}^{N_K \times 2}\), represented as \((x, y)\) coordinates of $N_K$ landmarks.

\begin{equation}
    \mathbf{p} = \text{UDP}(\arg\max_{\mathbf{p}'}(f(\mathbf{x})), f(\mathbf{x}))\label{eq1} .
\end{equation}


\subsubsection{PCT and YOLOv8}
Equation~\eqref{eq2} represents landmark detection using YOLOv8 and PCT. \(g\) is the landmark detection model. The output \(g(\mathbf{x}) \in \mathbb{R}^{N_K \times 2}\) directly predicts the coordinates of the \(N_K\) landmarks.
\begin{equation}
    \mathbf{p} = g(\mathbf{x})\label{eq2} .
\end{equation}

\subsection{Proposed Shape Subspace Refinement}\label{poserefinement}
We propose Shape Subspace Refinement (SSR) to refine landmark positions in heatmap-based landmark detection. While heatmap-based models are trained to produce a single peak on each heatmap, they may occasionally generate multiple or unclear peaks. 
In such cases, measuring the geometrical similarities between detected and reference landmark positions using shape subspace can help select the most likely landmark positions.
Fig.~\ref{fig_shape} provides a conceptual illustration of SSR. 
The shape subspaces for training and test samples $S_{train,i}$, $S_{test,c_j}$ are computed from their respective shape matrices $X_{train,i}$, $X_{test,c_j}$, which consist of landmark coordinates and are referred to as a point set.
Then, the similarity between the two subspaces is computed.
 
The key advantage of using shape subspace is its ability to represent the spatial relationships of a point set while remaining invariant under affine transformations applied to the point set.
This means that the shape subspace remains unchanged in case of positional shifts of the elbow bones in the images or size differences of the bone structures. Meanwhile, a non-linear transformation of the point set, such as when the position of a point in the point set is shifted, alters the shape subspace. 
Although the order of the points in the shape matrices must be consistent \cite{Igarashi}, our approach always satisfies this condition because each point coordinate is obtained from each heatmap.

\subsubsection{Shape Subspace}
The mathematical definition of shape subspace is provided below. 
A shape subspace \( \mathbf{S} \) is calculated from a shape matrix \( \mathbf{X} \) that contains 3D coordinates of an object:

\begin{equation}
    \mathbf{X} = (\mathbf{r}_1 \mathbf{r}_2 \dots \mathbf{r}_K)^T = 
    \begin{pmatrix}
        x_1 & y_1 & z_1 \\
        x_2 & y_2 & z_2 \\
        \vdots & \vdots & \vdots \\
        x_K & y_K & z_K
    \end{pmatrix},
\end{equation}

where \( \mathbf{r}_i = (x_i \; y_i \; z_i)^T \) for \( 1 \leq i \leq K \) denotes the positional vector of the \( i \)--th point on an object.

Next, the shape matrix \( \mathbf{X} \) is centered, and \( \mathbf{X}_c \) is obtained to satisfy the relation \( \sum_{i=1}^{K} \mathbf{r}_i = 0 \), making the shape subspace invariant to affine translation of the coordinates.

\begin{equation}
    \mathbf{X}_c = \mathbf{X} - \frac{1}{K} \sum_{i=1}^{K} \mathbf{r}_i.
\end{equation}

Then, Singular Value Decomposition (SVD) of the centered matrix \( \mathbf{X}_c \) is computed, and left singular vectors \( \mathbf{U} \), singular values \( \mathbf{\Sigma} \), and right singular vectors \( \mathbf{V}^T \) are obtained. The basis vectors of shape subspace \( \mathbf{S} \) are defined by the first \( 3 \) columns of \( \mathbf{U} \) (Algorithm~\ref{ssb}).

\begin{equation}
    \mathbf{X}_c = \mathbf{U} \mathbf{\Sigma} \mathbf{V}^T.
\end{equation}

The similarity between two shape subspaces \( \text{sim}(\mathbf{S}_1, \mathbf{S}_2) \) can be measured by the canonical angles \( \theta_i \), where \( \{ 0 \leq \theta_1, \dots, \theta_{N} \leq \frac{\pi}{2} \} \), with $N$ representing the subspace dimension. In this case, \( N = 3 \).

\begin{equation}
    \text{sim}(\mathbf{S}_1, \mathbf{S}_2) = \frac{1}{N} \sum_{i=1}^{N} \cos^2 \theta_i.
\end{equation}

Let \( \{ \mathbf{\Phi}_i \}_{i=1}^{N} \) be the basis vectors of \( \mathbf{S}_1 \) and \( \{ \mathbf{\Psi}_i \}_{i=1}^{N} \) be the basis vectors of \( \mathbf{S}_2 \). The projection matrices \( \mathbf{Q}_1 \) and \( \mathbf{Q}_2 \) are calculated as:

\begin{equation}
    \mathbf{Q}_1 = \sum_{i=1}^{N} \mathbf{\Phi}_i \mathbf{\Phi}_i^T, \quad
    \mathbf{Q}_2 = \sum_{i=1}^{N} \mathbf{\Psi}_i \mathbf{\Psi}_i^T.
\end{equation}

\( \cos^2 \theta_i \) is the \( i \)-th largest eigenvalue of \( \mathbf{Q}_1 \mathbf{Q}_2 \) or \( \mathbf{Q}_2 \mathbf{Q}_1 \) (Algorithm \ref{sim}).

\begin{algorithm}[t]
\caption{ShapeSubspaceBasis($X$, $N=3$)}
\KwIn{Shape matrix $X \in \mathbb{R}^{K \times N}$}
\KwOut{Basis vectors $S \in \mathbb{R}^{K \times N}$ of shape subspace}
$\mu \gets$ column-wise mean of $X$\;
$X_c \gets X - \mathbf{1}\mu^\top$\;
$[U, \Sigma, V^T] \gets \mathrm{SVD}(X_c)$\;
$S \gets U[:, :N]$\;
\Return{$S$}
\label{ssb}
\end{algorithm}

\begin{algorithm}[t]
\caption{ShapeSubspaceSimilarity($\Phi$, $\Psi$, $N=3$)}
\label{alg:subspace_similarity_single}
\KwIn{Subspace basis $\Phi \in \mathbb{R}^{K \times N}$ for $S_1$, subspace basis $\Psi \in \mathbb{R}^{K \times N}$ for $S_2$}
\KwOut{Similarity $\text{sim}(S_1, S_2) \in \mathbb{R}$}
$Q_1 \gets \Phi \Phi^\top$\;
$Q_2 \gets \Psi \Psi^\top$\;
$\text{sim} \gets \frac{1}{N} \mathrm{trace}(Q_1 Q_2)$\;
\Return{$\text{sim}$}
\label{sim}
\end{algorithm}

\begin{figure*}[h]
\centering
\includegraphics[width=0.95\textwidth]{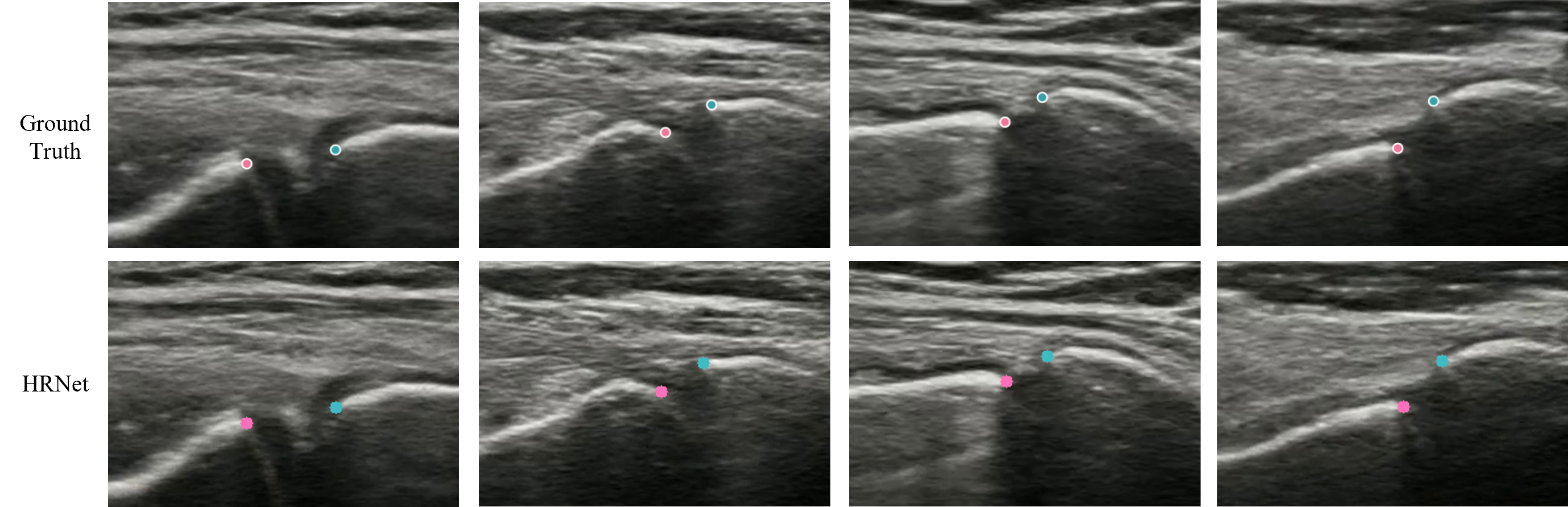}
\caption{The two-landmark detection results shown in enlarged views around the joints. The upper row displays the ground truth, and the lower row illustrates the inference results of HRNet. The pink dot is landmark1, and the light blue dot is landmark2.}\label{result-2point}
\end{figure*}

\subsubsection{Refinement Algorithm}

All possible combinations \( \mathit{c_j} \) of the selected landmark candidates are evaluated based on their similarities to the training data using shape subspace. Here, \( \mathit{c_j} \) represents a combination of \( (x, y) \) coordinates of \( \mathit{N_K} \) landmarks (i.e., the shape matrix $X_{\text{test},\,c_j}$), and \( \mathcal{C} \) denotes the set of all such combinations.
For each combination \( \mathit{c_j} \in \mathcal{C} \), its similarity to \( \mathit{M} \) training samples is computed, and the mean similarity is taken. Finally, the combination \( \mathit{c^*} \) that has the highest mean similarity is selected as follows:

\begin{equation}\label{shape_ave_max}
    \mathit{c^*} = \arg\max_{\mathit{c_j} \in \mathcal{C}} \left( \frac{1}{\mathit{M}} \sum_{i=1}^{\mathit{M}} \text{sim}(\mathbf{S}_{\text{train}, i}, \mathbf{S}_{\text{test}, \mathit{c_j}}) \right).
\end{equation}

The coordinates of the selected combination \( \mathit{c^*} \) are then input into the UDP algorithm along with the heatmaps \( f(\mathbf{x}) \), and the refined landmark positions \( \mathbf{p} \) are obtained.

\subsection{Point-based Bone Segmentation}\label{Segmentation}
The segmentation of the humerus and ulna using SAM2 is demonstrated using the detected landmarks as the point prompts.


\section{Experiments}\label{experiments}
In this section, landmark detection was performed on our proposed dataset to evaluate the accuracy of joint space measurement.
Model performance was assessed with two different landmark configurations: two-landmark detection (detecting the joint measurement points, landmarks 1 and 2) and eight-landmark detection (detecting all landmarks from 1 to 8).

\subsection{Experimental Configuration}\label{Configuration}
\subsubsection{Landmark Detection Models}
\begin{itemize}
{
    \item{ViTPose}\\
    The input image size was \(512 \times 384\) pixels, and the output heatmap size was \(128 \times 96\) pixels. The augmentation process included random scaling and rotation. The Normalized Mean Error (NME) was used as the evaluation metric. The total epoch was 210, with the best epoch being selected using the validation set. The optimizer was Adam. The initial learning rate was \(1 \times 10^{-4}\), and it was adjusted using a step-based policy, with reductions at epochs 170 and 200. The backbone was ViT-small, and the head was the classic decoder. The pre-trained model trained on ImageNet dataset was used. The patch size was \(16 \times 16\) pixels. The joint weights of all the landmarks were set to 1.0.
        
    \item{HRNet}\\
    The input image size was set to $512 \times 384$ pixels, and the output heatmap size was set to $128 \times 96$ pixels. The augmentation process included random scaling and rotation. The NME was used as the evaluation metric. The total epoch was 210, with the best epoch being selected using the validation set. The optimizer was Adam. The initial learning rate was \(1 \times 10^{-4}\), and it was adjusted using a step-based policy, with reductions at epochs 170 and 200. The backbone was HRNet-w32. The pre-trained model trained on ImageNet dataset was used. The joint weights of all the landmarks were set to 1.0. 
    
    \item{U-Net}\\
    The input image size and the output heatmap size were set to \(512 \times 512\) pixels. The augmentation process included random scaling and rotation. The normalized mean error was used as the evaluation metric. The total epoch was 600, with the best epoch selected by the validation set, and the learning rate was set to \(1 \times 10^{-3}\).
        
    \item{PCT}\\
    The input image size was set to $512 \times 512$ pixels. Data augmentation included random scaling, rotation, and shift. The Swin Transformer V2 backbone pre-trained with SimMIM on ImageNet was used. The optimizer was AdamW, with an initial learning rate of \(8 \times 10^{-4}\), which was adjusted using a cosine annealing schedule. The number of tokens was set to 34. The training epoch was set to 50 for the tokenizer and 210 for the classifier. The joint weights of all the landmarks were set to 1.0. These parameters were set following the official implementation \cite{bib_pct}.
    
    \item{YOLOv8}\\
    The input image size was \(640 \times 640\) pixels. The augmentation process included random rotations, scaling, shifts, and horizontal flips. Training was conducted for 1,000 epochs, with the best epoch selected for evaluation. The optimizer was AdamW for the first \(10,000\) iterations, followed by SGD. The initial learning rate was \(1 \times 10^{-3}\) for AdamW and \(1 \times 10^{-2}\) for SGD. The pose loss weight was set to 12.0.\\
}
\end{itemize}

\begin{figure*}[t!]
\centering
\includegraphics[width=0.9\textwidth]{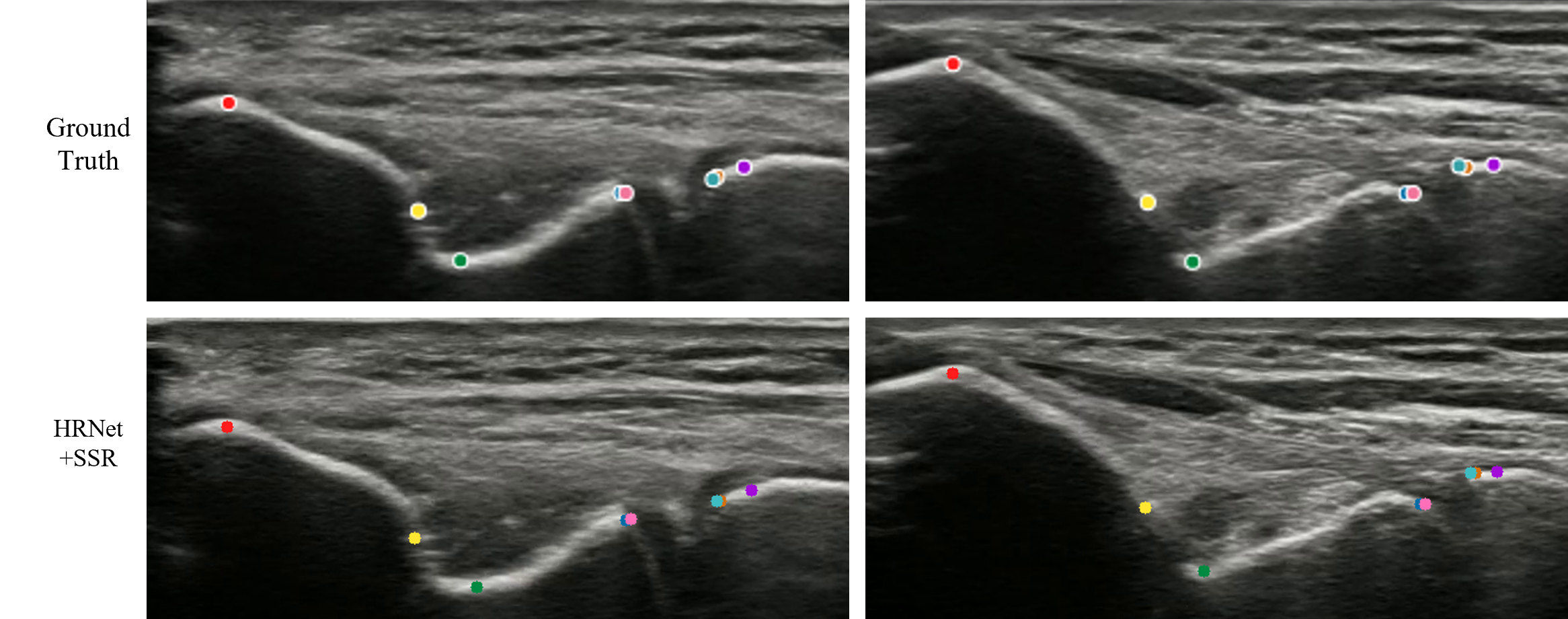}
\caption{The eight-landmark detection results. The first row shows the ground truth for 8 landmarks. The second row illustrates the positions of detected landmarks by HRNet+SSR.  The colors of landmarks 1 to 8 are pink, light blue, red, yellow, green, blue, orange, and purple.}\label{result-8point}
\end{figure*}

\subsubsection{Number of Landmarks}
Two-landmark and eight-landmark detection are compared. The two-landmark detection model detects landmarks 1 and 2. The eight-landmark model detects all landmarks from 1 to 8.
The landmark refinement is only applied to the eight-landmark detection.

\subsubsection{Parameter Settings for SSR}
For the landmark refinement, the value threshold $r$ was set to 75\%, and the length threshold $D$ was set to 5 pixels on the heatmap for the candidate selection. To calculate the similarity, random \( 3.3\% \) of the total training samples was used, which, for instance, corresponds to \( M = 67 \). 

\subsubsection{Point Prompts for the Point-based Segmentation}
We used the detected 6 points in the eight-landmark detection as the positive point prompts $(x_i, y_i),\ (i=3,..,8)$. Also, six negative point prompts based on the detected landmark coordinates $(x_4, y_3), (x_4, y_3),\ (x_5, y_4),\ (x_5, 2y_5 - y_4),\ (x_8, 2y_5 - y_4), \ ((x_1 + x_2)/2,\ (y_1 + y_2)/2)$ are used.

\begin{table}[h]
\caption{\textbf{The two-landmark detection results.} LM1 and LM2 represent the Mean Absolute Error (MAE) (mm) for landmarks 1 and 2, respectively. Ave2 is the average MAE of the two landmarks. Length denotes the mean Euclidean distance error (mm) for joint space length. The \textbf{bold values} indicate the lowest error in each column. \textbf{Smaller values indicate better detection performance.}}
\label{tab_2point}
\centering
\begin{tabular}{|l|cc|cc|}
\hline
& \multicolumn{2}{c|}{Joint} & \multicolumn{2}{c|}{} \\ 
\cline{2-3} 
\raisebox{1.5ex}[0pt]{Method} & LM1$\downarrow$ & LM2$\downarrow$ & \multicolumn{2}{c|}{\raisebox{1.5ex}[0pt]{Ave2$\downarrow$ \quad Length$\downarrow$}} \\ 
\hline
YOLOv8 & 0.793 & 1.753 & 1.340 & 1.276 \\
PCT & 0.579 & 0.762 & 0.848 & 0.730 \\
U-Net  &  0.239 & 0.150 & 0.194 & 0.212 \\
ViTPose & 0.168 & 0.161 & 0.164 & 0.172 \\
HRNet & \textbf{0.132} & \textbf{0.129} & \textbf{0.131} & \textbf{0.125} \\
\hline

\end{tabular}
\end{table}

\begin{table*}[h]
\caption{\textbf{The eight-landmark detection results.} LM1-LM8 represent Mean Absolute Error (MAE) (mm) for landmarks 1 through 8. Ave8 is the average MAE for all 8 landmarks (LM1--LM8), and Ave2 is the average MAE for LM1 and LM2. Length denotes the mean Euclidean distance error (mm) for joint space. SSR is Shape Subspace Refinement. The \textbf{bold values} indicate the lowest error in each column. \textbf{Smaller values indicate better detection performance.}}
\label{tab_8point}
\centering
\begin{tabular}{|l|cc|cccc|cc|ccc|}
\hline
& \multicolumn{2}{c|}{Joint} & \multicolumn{4}{c|}{Humerus}  & \multicolumn{2}{c|}{Ulna} & \multicolumn{3}{c|}{} \\ 
\cline{2-9} 
Method & LM1$\downarrow$ & LM2$\downarrow$ & LM3$\downarrow$ & LM4$\downarrow$ & LM5$\downarrow$ & LM6$\downarrow$ & LM7$\downarrow$ & LM8$\downarrow$ & \multicolumn{3}{c|}{\raisebox{1.5ex}[0pt]{Ave8$\downarrow$ \quad Ave2$\downarrow$ \quad Length$\downarrow$}} \\ 
\hline
YOLOv8                  & 0.568 & 0.664 & 0.949 & 0.584 & \textbf{0.871} & 0.571 & 0.646 & 1.127 & 0.706 & 0.616 & 0.574 \\
PCT                     & 0.624 & 0.833 & 0.919 & 0.711 & 1.301 & 0.637 & 0.841 & 0.903 & 0.860 & 0.728 & 0.708 \\
U-Net                   & 0.204 & 0.146 & 3.252 & 2.729 & 2.521 & 0.234 & 0.137 & 0.657 & 1.235 & 0.175 & 0.175 \\
U-Net + \textbf{SSR}    & 0.189 & \textbf{0.131} & 2.532 & 2.514 & 2.346 & 0.209 & \textbf{0.130} & 0.539 & 1.074 & \textbf{0.160} & \textbf{0.159} \\
ViTPose                 & 0.188 & 0.248 & 0.775 & 0.864 & 1.601 & 0.193 & 0.240 & 0.695 & 0.600 & 0.218 & 0.235 \\
ViTPose + \textbf{SSR}  & 0.170 & 0.227 & 0.612 & 0.757 & 1.030 & 0.183 & 0.207 & 0.526 & 0.464 & 0.199 & 0.217 \\
HRNet                   & \textbf{0.152} & 0.175 & 0.571 & 0.548 & 1.018 & \textbf{0.158} & 0.170 & 0.443 & 0.404 & 0.164 & 0.169 \\
HRNet + \textbf{SSR}    & \textbf{0.152} & 0.175 & \textbf{0.515} & \textbf{0.536} & 0.891 & \textbf{0.158} & 0.169 & \textbf{0.442} & \textbf{0.380} & 0.163 & 0.169 \\

\hline

\hline
\end{tabular}
\end{table*}

\begin{table*}[h]
\centering
\caption{\textbf{Fold-wise two- and eight-landmark detection results.}
Ave2 is the average MAE for LM1 and LM2, and Ave8 is the average MAE for all 8 landmarks (LM1–LM8). Length denotes the mean Euclidean distance error for joint space.
SSR is Shape Subspace Refinement.
All values are shown as mean ± standard deviation (in mm). \textbf{Smaller values indicate better detection performance.}}

\begin{tabular}{|l|cc|ccc|}
\hline
& \multicolumn{2}{c|}{\textbf{Two-landmark detection}} & \multicolumn{3}{c|}{\textbf{Eight-landmark detection}} \\
& \multicolumn{2}{c|}{\textbf{HRNet}} & \multicolumn{3}{c|}{\textbf{HRNet+SSR}} \\
\textbf{Fold} & \textbf{Ave2$\downarrow$} & \textbf{Length$\downarrow$} & \textbf{Ave8$\downarrow$} & \textbf{Ave2$\downarrow$} & \textbf{Length$\downarrow$} \\
\hline
fold 1-1 &      0.141 ± 0.141 &  0.153 ± 0.182 &    0.460 ± 1.118 & 0.196 ± 0.129 & 0.269 ± 0.228 \\
fold 1-2 &      0.147 ± 0.106 &  0.106 ± 0.148 &    0.338 ± 0.383 & 0.188 ± 0.115 & 0.127 ± 0.158 \\
fold 1-3 &      0.158 ± 0.308 &  0.195 ± 0.442 &    0.420 ± 0.686 & 0.188 ± 0.150 & 0.208 ± 0.263 \\
fold 1-4 &      0.125 ± 0.077 &  0.076 ± 0.092 &    0.429 ± 1.125 & 0.140 ± 0.083 & 0.136 ± 0.136 \\
fold 1-5 &      0.130 ± 0.091 &  0.127 ± 0.143 &    0.361 ± 0.623 & 0.165 ± 0.091 & 0.155 ± 0.193 \\
\hline
fold 2-1 &      0.118 ± 0.081 &  0.127 ± 0.159 &    0.341 ± 0.528 & 0.150 ± 0.150 & 0.169 ± 0.176 \\
fold 2-2 &      0.125 ± 0.141 &  0.113 ± 0.183 &    0.421 ± 1.382 & 0.158 ± 0.158 & 0.166 ± 0.436 \\
fold 2-3 &      0.127 ± 0.478 &  0.130 ± 0.426 &    0.286 ± 0.276 & 0.160 ± 0.160 & 0.182 ± 0.218 \\
fold 2-4 &      0.128 ± 0.078 &  0.138 ± 0.166 &    0.415 ± 0.690 & 0.162 ± 0.162 & 0.175 ± 0.208 \\
fold 2-5 &      0.111 ± 0.081 &  0.083 ± 0.106 &    0.326 ± 0.395 & 0.126 ± 0.126 & 0.101 ± 0.121 \\
\hline
\end{tabular}
\label{tab:fold-wise-results}
\end{table*}

\subsection{Evaluation Metrics}
Mean Absolute Error (MAE) was used to evaluate the accuracy of each predicted landmark.
\begin{equation}
    \text{MAE}_k = \frac{1}{N_t} \sum_{i=1}^{N_t} {\sqrt{(x_{i,k} - x_{i,k}^{\text{gt}})^2 + (y_{i,k} - y_{i,k}^{\text{gt}})^2}} .
\end{equation}

where \(N_t\) represents the total number of test samples, \(x_{i,k}\) and \(y_{i,k}\) are the predicted \(x\)- and \(y\)-coordinates of the \(k\)-th landmark in the \(i\)-th image, and \(x_{i,k}^{\text{gt}}\) and \(y_{i,k}^{\text{gt}}\) are the ground truth coordinates of the same landmark.

Also, Euclidean Distance Error (EDE) was used to evaluate the joint space length. The EDE is defined as the mean Euclidean distance error across all $N_t$ samples.

\[
\text{EDE} = \frac{1}{N_t} \sum_{i=1}^{N_t} \left| {\|\mathbf{p_{1,i}} - \mathbf{p_{2,i}}\|}_2 - {\|\mathbf{p_{1,i}^{\text{gt}}} - \mathbf{p_{2,i}^{\text{gt}}}\|}_2 \right|.
\]

where \({\|\mathbf{p_{1,i}} - \mathbf{p_{2,i}}\|}_2\), \({\|\mathbf{p_{1,i}^{\text{gt}}} - \mathbf{p_{2,i}^{\text{gt}}}\|}_2\) represent the Euclidean distances for the predicted and ground truth lengths between landmarks 1 and 2, respectively.

Our dataset of 22 participants was split into 5 groups based on individuals. We performed 5-fold cross-validation under two different split settings. In each fold, 3 groups were used for training, 1 for validation, and 1 for testing; for example, one such split included 13 participants for training, 4 for validation, and 5 for testing. The full-size images without cropping were used in the experiment. The final results were averaged over all ten folds across the two split settings. The samples included in each split are detailed in the supplementary material.

\section{Results}
This section presents the results of the two-landmark and eight-landmark detection and point-based segmentation.

\subsection{The Two-Landmark Detection}\label{2point}
\begin{figure*}[h]
\centering
\includegraphics[width=0.9\textwidth]{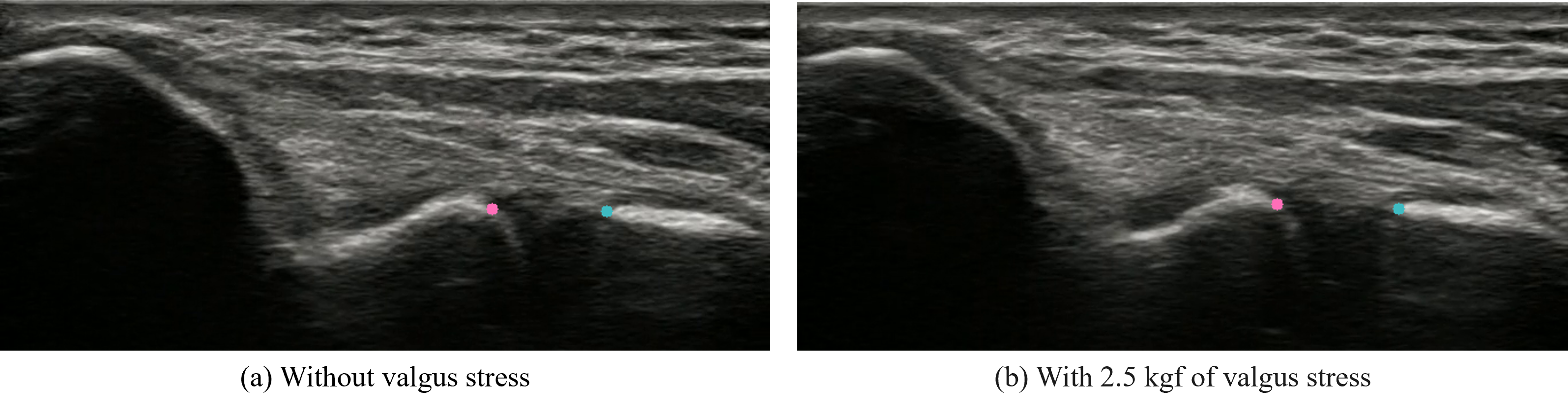}
\caption{A pair of inference results on the medial elbow of a baseball player using our approach:  
(a) without stress,
(b) with 2.5  kgf of applied valgus stress.
}\label{withwithout_stress}
\end{figure*}

Table~\ref{tab_2point} presents the results of two-landmark detection, showing that heatmap-based methods outperformed other approaches. Among them, HRNet achieved the best performance, with an average MAE of $0.131$ for the two landmarks and an EDE of $0.125$ for the joint space.
As illustrated in Fig.~\ref{result-2point}, the positions of landmarks 1 and 2 are precisely detected, even in the presence of variations in bone shapes and white noise in the joint space.
Fold-wise results for HRNet are detailed in Table~\ref{tab:fold-wise-results}.
To further demonstrate the applicability of our method to diagnose UCL injury, joint space measurements were conducted with and without 2.5 kgf of valgus stress applied proximally to the wrist.
Fig.~\ref{withwithout_stress} shows HRNet-based inference results on the medial elbow of a baseball player under both stressed and unstressed conditions, demonstrating accurate detection in both cases. Additional results from eight baseball players not included in the dataset are provided in the supplementary material.


\subsection{The Eight-Landmark Detection}\label{8point}
Table ~\ref{tab_8point} presents the results for the eight-landmark detection. 
Among the conditions, HRNet with SSR showed the best performance in the average MAE (Ave8, Ave2). 
Also, the average MAE using SSR was always equal to or less than that without SSR, indicating that shape subspace effectively refines landmark locations.
Fig.~\ref{result-8point} illustrates the detected landmarks in the eight-landmark detection using HRNet with SSR, which demonstrates that the landmarks are localized accurately.
The fold-wise results for HRNet with SSR are summarized in Table~\ref{tab:fold-wise-results}.

\subsection{Point-based Segmentation}\label{pointseg}
\begin{figure*}[t!]
\centering
\includegraphics[width=0.95\textwidth]{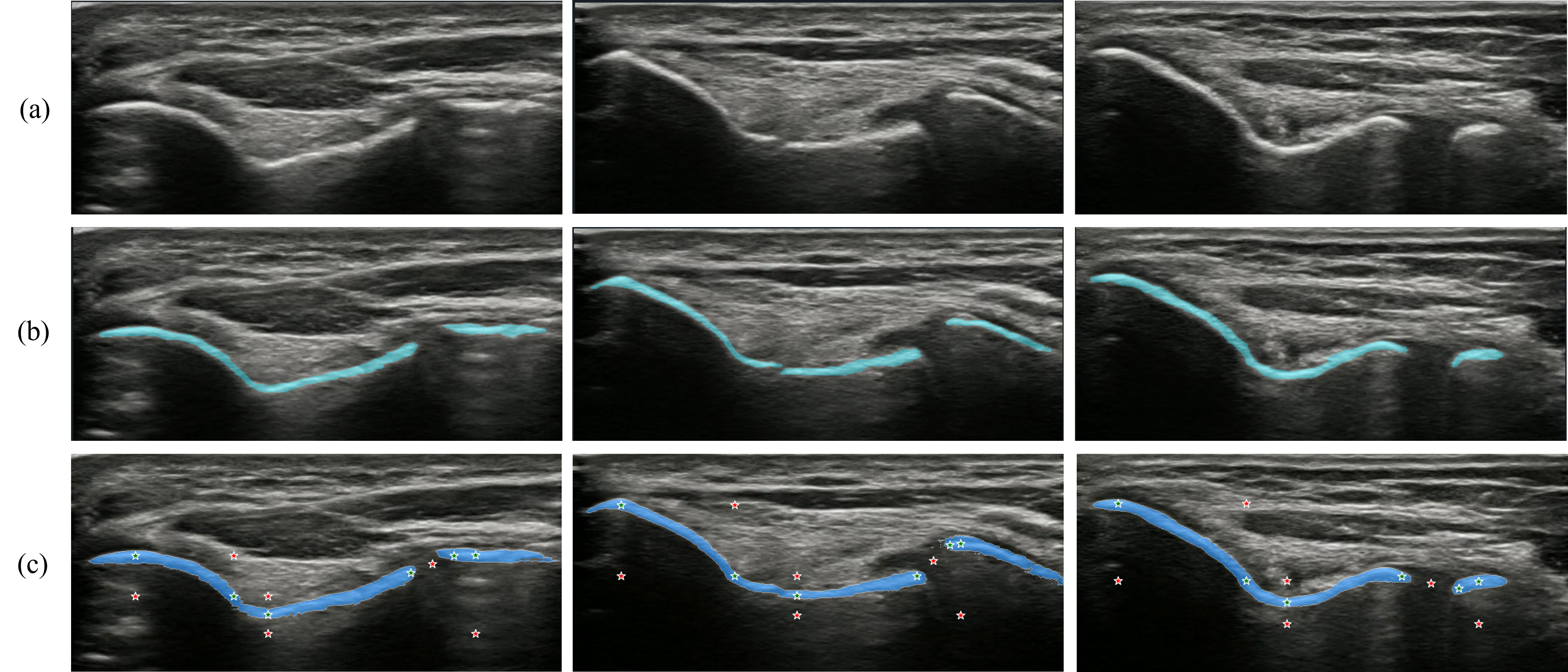}
\caption{The point-based segmentation results using SAM2 with point prompts derived from the detected landmark positions. (a) Original images. (b) Ground truth of the humerus and ulna, shown in blue-green. (c) Segmented images of the humerus and ulna using SAM2 with point prompts derived from detected landmarks. The blue areas indicate the segmented regions, green stars mark the positive point prompts, and red stars denote the negative point prompts. }\label{sam2}
\end{figure*}

Fig.~\ref{sam2}(c) presents the point-based segmentation results of SAM2, using the detected landmarks from HRNet with SSR as point prompts.
Green stars indicate positive prompts, red stars denote negative prompts, and blue areas represent segmented regions. This figure demonstrates that the humerus and ulna are segmented with high fidelity.

\begin{figure*}[t!]
\centering
\includegraphics[width=\textwidth]{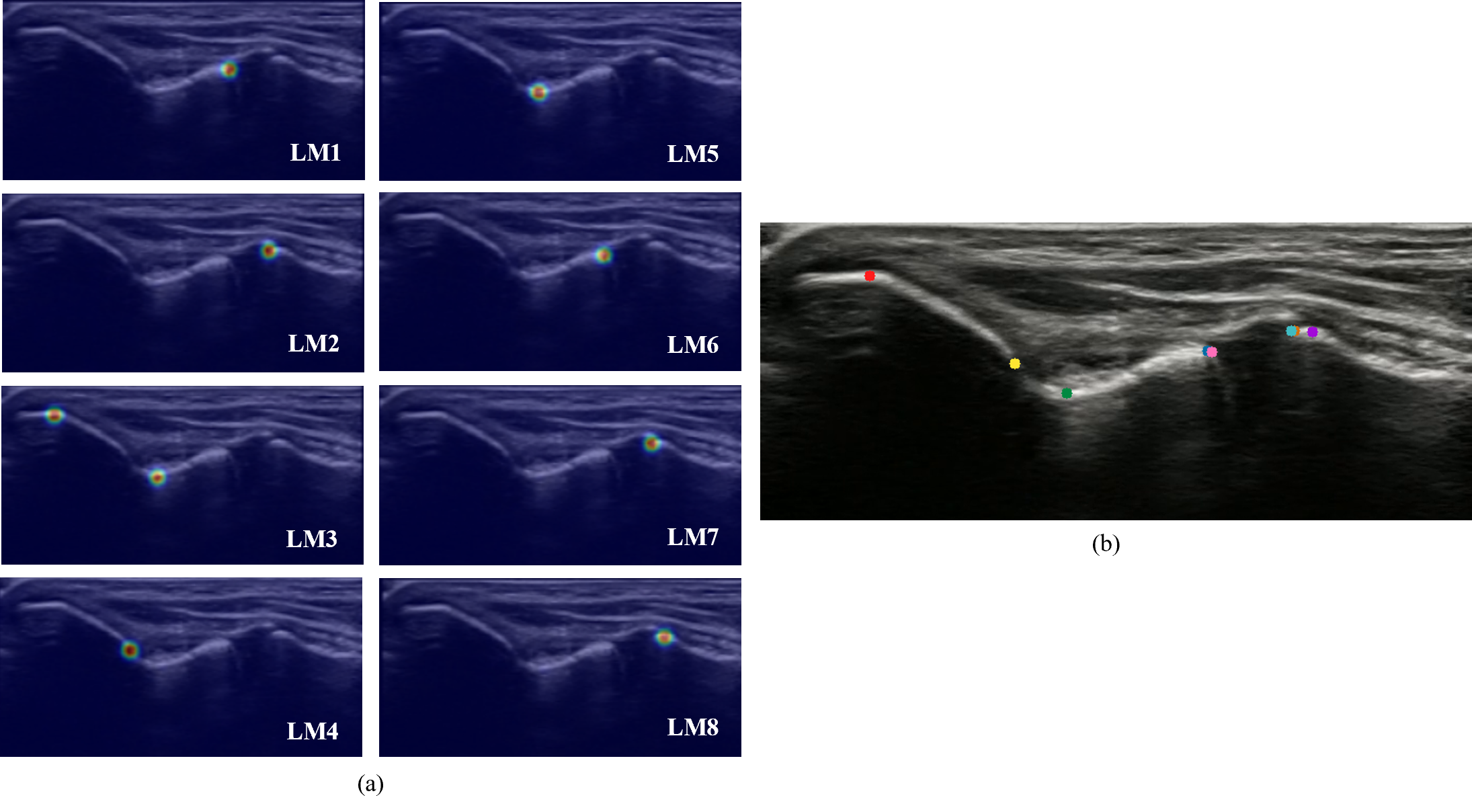}
\caption{An example of landmark refinement by SSR. (a) The overlays of the output heatmaps on the input image. Multiple peaks are observed in the LM3 heatmap. (b) The localized landmarks after applying SSR. The detected landmarks 1–8 are shown in pink, light blue, red, yellow, green, blue, orange, and purple, respectively.
In the LM3 heatmap, although the right peak exhibits higher intensity, SSR correctly selects the left peak, shown as the red point, which is closer to the ground truth.
}\label{multiple_peaks}
\end{figure*}

\section{Discussion}
First, the heatmap-based methods outperformed other approaches in both results. Among them, HRNet achieved the highest performance. The ability of HRNet to capture local features while maintaining high-resolution representations makes it particularly well-suited for precise landmark detection in ultrasound images.
In contrast, Transformer-based models like ViTPose typically require larger datasets for effective training. Given the limited dataset size, HRNet's efficiency with smaller datasets may have contributed to its superior performance.

Second, the two-landmark detection demonstrated better accuracy in joint space measurements than the eight-landmark detection model. This is likely because the simpler task of detecting fewer landmarks allows the model to focus on the most relevant features.

\begin{figure*}[h]
\centering
\includegraphics[width=0.9\textwidth]{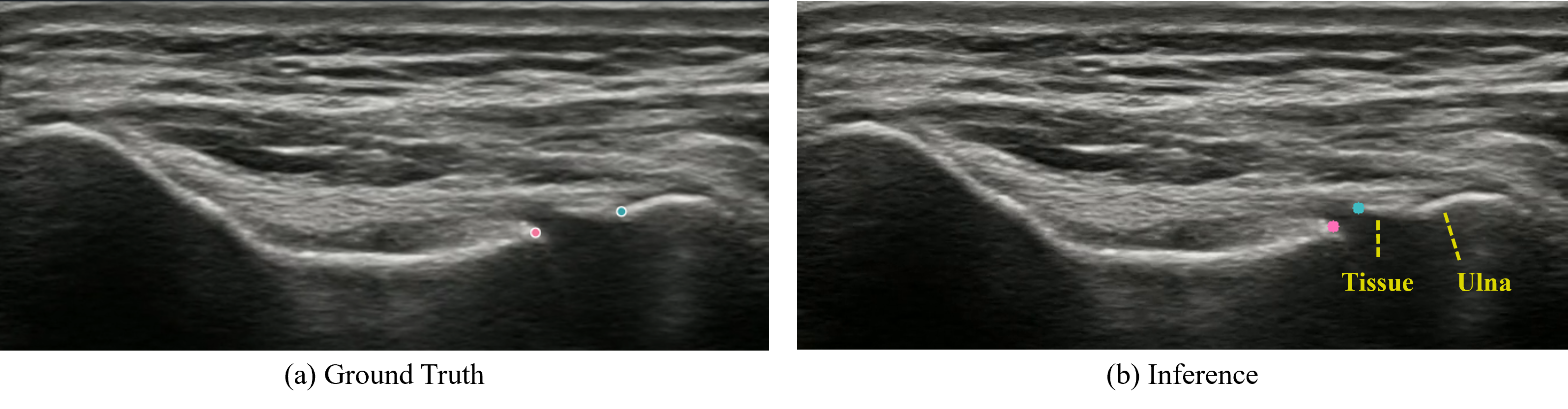}
\caption{A representative failure case in which the model predicted the proximal end of the ulnar coronoid process to the left of the ground truth, likely due to a tissue-like structure visually connected to the ulna’s left edge.
}\label{Failure}
\end{figure*}

Third, in the eight-landmark model, SSR effectively reduced landmark position errors.
The degree of error reduction achieved by SSR varied across models: 0.025 mm for HRNet, 0.136 mm for ViTPose, and 0.161 mm for U-Net. This variation is attributed to differences in multiple peak detection rates in the output heatmaps, which were 12.1\% for HRNet, 30.1\% for ViTPose, and 66.2\% for U-Net. 
These results indicate that SSR becomes more effective in improving landmark localization accuracy as the multiple peak detection rate increases.
Fig.~\ref{multiple_peaks} shows an SSR example, with the LM3 heatmap having multiple peaks.
In the LM3 heatmap, although the right peak has a higher value than the left peak, SSR correctly selects the left peak that aligns with the ground truth position of landmark 3, as shown by the red point in Fig.~\ref{multiple_peaks}(b).
The SSR inference required approximately 0.0006 seconds per image, and HRNet took 0.0128 seconds, resulting in a total of about 0.0134 seconds per image, making it suitable for real-time applications. This measurement was performed on a machine with an AMD Ryzen 9 5950X CPU and a single NVIDIA RTX 3090 GPU.
Such rapid inference aligns with the broader trend of employing AI-based diagnosis \cite{ai-diagnosis0}, \cite{ai-diagnosis1}.

Finally, we evaluated whether our method meets the precision requirement for the diagnosis of UCL injuries.
Pitchers with UCL injuries typically exhibit an average joint space increase of 1.2~mm under stressed conditions~\cite{bib_gap1}. In comparison, our method achieved an average error of 0.125~mm using HRNet in two-landmark detection, which is substantially smaller than this pathological threshold. Also, to assess whether the 1.2~mm difference is distinguishable from measurement noise, we applied the 3-sigma rule. Let $\Delta$ denote the difference in joint space measurements between the stressed and unstressed conditions. Assuming equal variance under both conditions, the standard deviation of the difference is given by:
$
\sigma_\Delta = \sqrt{2} \, \sigma.
$
This yields a limit of detection (LoD) defined as:
$
\mathrm{LoD} = 3 \sqrt{2} \, \sigma.
$
Therefore, the measurement is considered distinguishable if:
$
\sigma \leq \frac{1.2}{3\sqrt{2}} \approx 0.283 ~\text{mm}.
$
In five-fold cross-validation, four out of five folds in each split exhibited standard deviations ranging from 0.092 to 0.183 that are below the threshold 0.283~mm, while only one fold per split showed a higher deviation (up to 0.442) as shown in Table~\ref{tab:fold-wise-results} (see Length values under Two-landmark detection).
These results indicate that our method generally yields stable measurements, with increased variability observed only in one fold per split. 
This variability may be attributed to artifacts or imaging ambiguity, such as vague edges of the ulna and the presence of tissue-like structures that appear visually connected to the edge of the ulna, as illustrated in Fig.~\ref{Failure}.

\subsection{Limitations and Future Work}
First, the dataset was acquired using specific ultrasound equipment, which may not fully capture clinical variability, such as differences in machines and probes. Second, despite the dataset comprising data from participants from diverse backgrounds, the model may have limited generalization to populations with different anatomies. To address these limitations, future work could include collecting data across various devices and anatomical variations such as humerus avulsion cases and pediatric elbows, followed by model retraining to enhance robustness and clinical applicability.
Third, while shape subspace refines landmarks when multiple peaks exist, it may struggle when peaks are closely connected over a few pixels or when no clear peak is present in the heatmap.

\section{Conclusion}
In this paper, we introduced a novel ultrasound dataset consisting of 4,109 medial elbow images from 22 individuals, annotated with landmarks under the supervision of orthopedic experts. Joint space measurement methods trained on our dataset were evaluated, with HRNet achieving a mean joint space measurement error of 0.125 mm, which demonstrated sufficient precision for detecting UCL injuries. In addition, we proposed a shape subspace refinement method, which effectively improved landmark localization performance.
These results suggest that our approach can facilitate the early diagnosis and intervention of UCL injuries by enabling automated joint space measurement.
Furthermore, the point-based bone segmentation of the humerus and ulna is demonstrated using detected landmarks as point prompts, highlighting a potential application of our dataset and methodology in musculoskeletal ultrasound analysis.

\section*{Acknowledgment}


The authors appreciate Prof. Kazuhiro Fukui, Mr. Matheus Silva de Lima, and Mr. Santos Enoque Safrao for their guidance on the shape subspace implementation. Special thanks are extended to Prof. Xie Chun for his continuous support and insightful advice throughout this work. 
The authors would like to thank Prof. Yasukazu Totoki for his support as the principal investigator of the JSPS KAKENHI project (Grant Number 24K14397). 
Finally, gratitude is expressed to all the participants for their cooperation and consent.

\EOD

\end{document}